%% file: bare_jrnl.tex
\documentclass[10pt,twocolumn,twoside]{IEEEtran}

\usepackage{times}
\usepackage{epsfig}
\usepackage{graphicx}
\usepackage{amsmath}
\usepackage{amssymb}
\usepackage{url}
\usepackage{makecell}
\usepackage{multirow}
\usepackage[table,xcdraw]{xcolor}
\usepackage{subfigure}
\usepackage{threeparttable}
\usepackage{bbold}
\usepackage{dsfont}
\usepackage{bbm}
\usepackage[numbers,sort&compress]{natbib}
\usepackage{array}
\usepackage{booktabs}

\usepackage{expl3}
\ExplSyntaxOn
\newcommand\latinabbrev[1]{
	\peek_meaning:NTF . {
		#1\@}%
	{ \peek_catcode:NTF a {
			#1.\@ }%
		{#1.\@}}}
\ExplSyntaxOff

\def\ie{\latinabbrev{i.e}}

\def\ie{\latinabbrev{i.e}}

\newlength\secmargin
\newlength\subsecmargin
\newlength\paramargin
\newlength\figmargin
\newlength\eqmargin
\ifCLASSINFOpdf
\else
\fi

\begin{document}

\title{Learning Multi-expert Distribution Calibration for Long-tailed Video Classification}

\author{Yufan~Hu,
	Junyu~Gao,
	and~Changsheng~Xu,~\IEEEmembership{Fellow,~IEEE}
	\thanks{Yufan Hu is with University of Science and Technology Beijing, Beijing 100083, China, and also with Peng Cheng Laboratory, ShenZhen 518055, China. (e-mail: huyufanqaixuan@gmail.com).}
	\thanks{Junyu Gao and Changsheng Xu are with National Lab of Pattern Recognition, Institute of Automation, Chinese Academy of Sciences, Beijing 100190, P. R. China, and with School of Artificial Intelligence, University of Chinese Academy of Sciences, Beijing, China. Changsheng Xu is also with Peng Cheng Laboratory, ShenZhen 518055, China. (e-mail: \{junyu.gao, csxu\}@nlpr.ia.ac.cn).}
	\thanks{Copyright (c) 2020 IEEE. Personal use of this material is permitted. However, permission to use this material for any other purposes must be obtained from the IEEE by sending a request to pubs-permissions@ieee.org.}}

%
%



\markboth{IEEE Transactions on Multimedia,~Vol. XX, ~No. XX,~Aug 201X}
{Hu \MakeLowercase{\textit{et al.}}: Learning Multi-experts Distribution Calibration for Long-tailed Video Classification}

%


\maketitle
\begin{abstract}
	\input{abst.tex}
\end{abstract}

\begin{IEEEkeywords}
long-tailed distribution, video classification, multi-expert calibration.
\end{IEEEkeywords}

%
\IEEEpeerreviewmaketitle

\vspace{\secmargin}
\section{Introduction}
\vspace{\secmargin}
\input{intro.tex}

\vspace{\secmargin}
\section{Related Work}\label{sec:related_work}
\vspace{\secmargin}
\input{rw.tex}

\vspace{\secmargin}
\section{Our Approach}\label{sec:our}
\vspace{\secmargin}
\input{our.tex}

\vspace{\secmargin}
\section{Experimental Results}\label{sec:experimental_results}
\vspace{\secmargin}
\input{expr.tex}

\vspace{\secmargin}
\section{Conclusion}\label{sec:conclusion}
\vspace{\secmargin}
\input{conclus.tex}





%

%
%
%
%
%

\ifCLASSOPTIONcaptionsoff
\newpage
\fi



%
{
	\small
	\bibliographystyle{IEEEtran}
	\bibliography{ref}
}

\end{document}

%% file: abst.tex
Most existing state-of-the-art video classification methods assume that the training data obey a uniform distribution. However, video data in the real world typically exhibit an imbalanced long-tailed class distribution, resulting in a model bias towards head class and relatively low performance on tail class. While the current long-tailed classification methods usually focus on image classification, adapting it to video data is not a trivial extension. We propose an end-to-end multi-expert distribution calibration method to address these challenges based on two-level distribution information. The method jointly considers the distribution of samples in each class (intra-class distribution) and the overall distribution of diverse data (inter-class distribution) to solve the issue of imbalanced data under long-tailed distribution. By modeling the two-level distribution information, the model can jointly consider the head classes and the tail classes and significantly transfer the knowledge from the head classes to improve the performance of the tail classes. Extensive experiments verify that our method achieves state-of-the-art performance on the long-tailed video classification task.

%% file: intro.tex
\begin{figure}[t]
	\centering
	\includegraphics[scale=0.48]{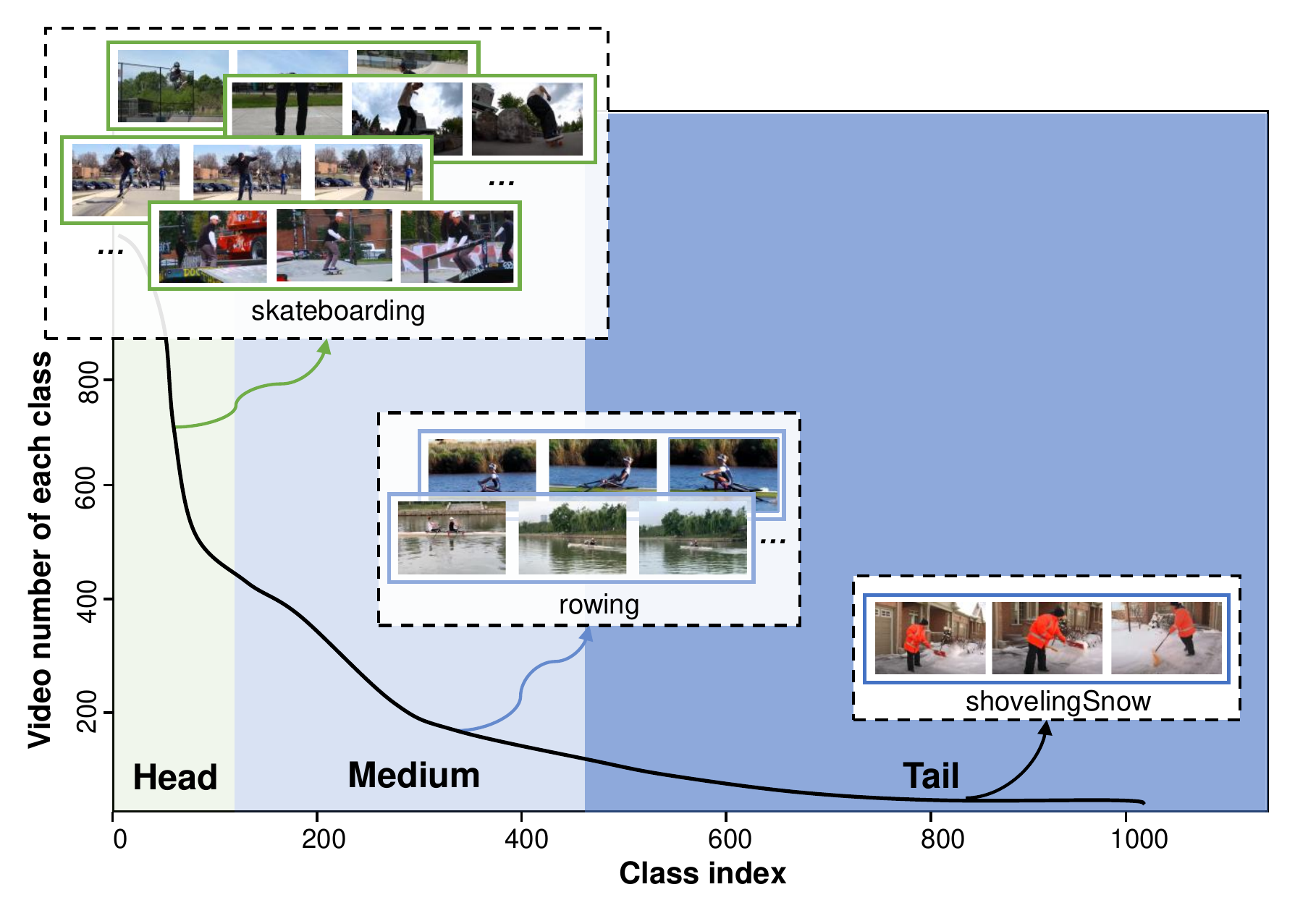}
	\caption{Real-world datasets exhibit long-tailed distribution. For head or medium classes, like ``skateboarding'' and ``rowing'', the large amount of samples lead to model bais and overfitting problem. When faced with tail classes, like a ``shovelingSnow'', the extreme imbalance and lack of samples cause tremendous challenges on the classification accuracy.
	} 
	\label{fig:1}
\end{figure}

Video classification has a wide range of applications in different scenarios such as video surveillance \cite{nascimento2006performance,sultani2018real}, self-driving cars \cite{kataoka2018temporal}, health-care \cite{dentamaro2019real}, \emph{etc}. With the advent of deep Convolutional Neural Networks, the performance of video classification has witnessed incredible progress \cite{tran2015learning,feichtenhofer2019slowfast,wang2021tdn,jiang2018modeling}. A series of CNN models~\cite{simonyan2014two,yue2015beyond,karpathy2014large} focus on video classification learning, which require access to abundant labels and uniformly distributed classes. However, as shown in Figure \ref{fig:1}, video data often exhibit extreme long-tailed (LT) distribution in the real world, where a few common classes (head classes) contain plenty of samples, and limited samples can underrepresent some rare categories (tail classes). Due to the naturally imbalanced distribution of training data, the trained model would be easily biased towards head classes and performs poorly on tail classes. To deal with this issue, recent trends in video classification have led to a proliferation of studies that train models with the long-tailed distribution dataset. For example, Zhang \textit{et al.} \cite{zhang2021videolt} propose the first large-scale and long-tailed untrimmed video dataset and perform evaluation against currently popular visual LT-recognition approaches.

To tackle the severe data imbalance issue in LT classification, typical methods can be roughly divided into two categories, re-sampling and re-weighting. Re-sampling is a regular strategy for imbalanced datasets \cite{chawla2002smote,wallace2011class,buda2018systematic}, where the basic idea is to over-sample tail classes and under-sample head classes. Re-weighting \cite{menon2013statistical,cui2019class,ren2018learning} is another commonly used technique to balance imbalanced data by up-weighting the tailed samples and down-weighting the head samples in the loss function. Currently, these methods mainly focus on the image classification tasks while few efforts have been made for the video domain. Compared with static images, a video is generally composed of hundreds of frames containing various scene dynamics. It has more complex structures than images, involving temporal information and more noises. Therefore, long-tailed video classification can be more difficult to understand the concept in a video of tail classes when only few examples are provided. Although the models proposed or adopted by~\cite{zhang2021videolt} show favorable performance, they still belong to the traditional re-sampling or re-weighting methods and suffer from the following problems:
\begin{figure*}[t]
	\centering
	\includegraphics[scale=0.55]{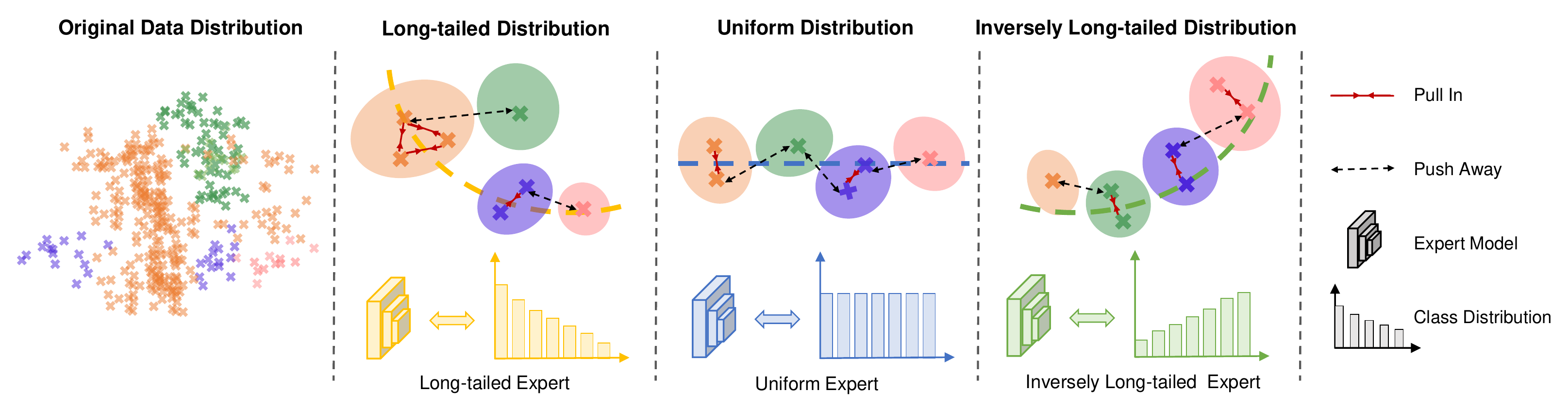}
	\caption{The motivation of our multi-expert distribution calibration method. Each point from the original data distribution can be used to estimate the intra-class distributions. Different categories are marked with different colors. The solid red lines represent that samples from the same class are pushed in, and the dashed black lines represent that different classes are pushed away. To handle various data distribution in the real world, we use three experts for model training including long-tailed, uniform, and inversely long-tailed experts. }
	\label{fig:2}
\end{figure*}

\begin{itemize}
	\item \textit{1)}: Most existing long-tailed classification methods only model a sample during training while neglecting the relations between samples and the distributions of samples from the same or different classes. When the training data are sufficient, the classifier for a specific class can be estimated accurately. However, the classifier is prone to overfit due to the scarcity of samples in the tail classes. In fact, the samples in each class exhibit a specific distribution. If such class distribution can be estimated, the representation of each category can be better modeled to effectively deal with the under-learning problem caused by insufficient data.

	\item \textit{2)}: In long-tailed classification, training classes exhibit long-tailed distribution while the distribution of test categories may be different. Most approaches assume that each test class has the same number of test samples, while the assumption does not always fit in real-world applications. The distribution of test classes may have various types of class distribution, such as uniform, long-tailed, or even inversely long-tailed distribution. It is necessary to consider the case where the test distribution is different from the training distribution. 
	
\end{itemize}

To address the above problems, we propose an end-to-end multi-expert distribution calibration (MEDC) approach by jointly considering two-level distributions, \emph{\ie}, intra-class distribution and inter-class distribution, to effectively solve the problem of data imbalance under long-tailed distribution. As shown in Figure \ref{fig:2}, the intra-class distribution represents the distribution between samples in a class and the inter-class distribution shows the overall distribution of diverse data. With the two-level distribution, the model can learn tail classes as well as head classes and significantly transfer the knowledge of the head classes to the tail classes. Specifically, for the intra-class distribution, we assume that the feature of samples in each class follows a Gaussian distribution in the latent space. Our model performs probabilistic modeling by estimating mean and variance, which are the parameters of the Gaussian distribution. The mean of the distribution can be regarded as the class prototype, while the variance represents the differences of samples in the same categories. Therefore, we can build a comprehensive relation among samples in each category to better model the head, medium, and tail classes. However, sampling operation in the distribution is non-derivative, we therefore propose to directly use samples to estimate the mean and variance of a class.



Furthermore, the intra-class distribution of each category constitutes the overall distribution of the data, \emph{\ie}, inter-class distribution. The real-world data may follow any inter-class distribution, such as uniform, long-tailed, or even inversely long-tailed distribution. For model training, given a particular type of inter-class distribution, the model has the ability to handle the test class with the same distribution. Therefore, when using different types of inter-class distributions to guide model learning, the model will have the ability to operate various distributions of test classes. Considering such different inter-class distributions enables the model to improve the generalization ability and learn feature representations of different categories. Motivated by such observation, we learn a multi-expert model from the original long-tailed training dataset, which can deal with various inter-class distributions.

In conclusion, we propose a unified multi-expert distribution calibration framework for long-tailed video classification, where multiple experts are trained to be skilled in the different inter-class distributions and the distribution calibration strategy is used to consider the intra-class distribution jointly. The main contributions of this paper can be summarized as follows:

\begin{itemize}
	\item We jointly consider the modeling of two-level distributions, including intra-class distribution and inter-class distribution, and verify the effectiveness of probabilistic estimation of the distributions in the long-tailed video classification task.
	
	\item We explore a multi-expert distribution calibration method for long-tailed video classification. MEDC uses a multi-expert mechanism to model diversified inter-class distributions and uses a distribution calibration strategy to adjust the relation between intra-class distributions in each expert. The two modules are co-trained under a unified framework and mutually reinforce each other.
	 
	\item We evaluate our model on a large-scale long-tailed video classification dataset VideoLT \cite{zhang2021videolt}, and our proposed model consistently achieves superior performance over previous state-of-the-art approaches.

\end{itemize}


%% file: rw.tex
This section reviews three topics closely related to our work in terms of deep learning-based video classification, long-tailed visual classification methods, and distribution learning.

\subsection{Deep Learning-based Video Classification}

Video is composed of continuous image sequences with complex motion information and naturally has spatial and temporal dimensions. Early studies on video classification directly utilize hand-crafted features to model classifiers \cite{niebles2008unsupervised,laptev2008learning,niebles2010modeling}. Hand-crafted features rely heavily on human experience or domain knowledge. The experience and knowledge may be useful in certain task-specific settings, while this will result in a longer cost time to develop a successful video classification method for more general environments and tasks.

Nowadays, deep learning has shown its ability to learn more discriminative and robust features \cite{xie2015hybrid,jaderberg2015spatial}. Deep learning abandons the paradigm of using artificially designed features directly and utilizes deep neural networks to learn features automatically. Besides, deep learning achieves great progress in video classification \cite{simonyan2014two,donahue2015long,wang2016temporal,gao2020unsupervised,hu2020learning,gao2020ci}. Video classification methods can be summarized into four categories, namely methods based on spatial-temporal networks, multiple stream networks, deep generative networks
and temporal coherency networks following \cite{herath2017going}. (1) Spatial-temporal networks \cite{yue2015beyond,ji20123d,tran2015learning,hu2022learning} use convolutional operation to extract features from both spatial and temporal domains. Ng \textit{et al.} \cite{yue2015beyond} develop convolutional temporal feature pooling architecture to aggregate spatial and temporal information and model the video as an ordered sequence of frames based on long short-term memory network (LSTM). Karpathy \textit{et al.} \cite{karpathy2014large} utilize local spatio-temporal information to extend CNNs into video classification on a large-scale dataset. (2) Multiple stream networks \cite{wu2016multi,feichtenhofer2016convolutional} combine the multi-modal information, such as frame, optical flow and audio as inputs of deep networks. Feichtenhofer \textit{et al.} \cite{feichtenhofer2016convolutional} take advantage of spatio-temporal information by fusing ConvNet towers both spatially and temporally. Wu \textit{et al.} \cite{wu2015fusing} learn to exploit multi-modal features, such as spatial, short-term motion and audio clues, using a multi-stream framework. (3) Deep generative networks \cite{yan2014modeling,srivastava2015unsupervised} adopt generative models to accurately learn the underlying distribution of data. Vincent \textit{et al.} \cite{vincent2008extracting} learn the temporal data in an unsupervised matter inspired from a manifold learning and information theoretic perspective. Goodfellow \textit{et al.} \cite{goodfellow2014generative} utilize backpropagation and dropout algorithms to train the generative and the discriminative model. (4) Temporal coherency networks \cite{goroshin2015unsupervised,rahmani2015learning} take the notion of temporal coherency into perspective, where the temporal coherency means that video frames are correlated both semantically and dynamically. Fernando \textit{et al.} \cite{fernando2015modeling} capture video-wide temporal information by utilizing the parameters of the ranking functions as a new video representation for video classification. Most existing video classification approaches assume that the data of each category is balanced. Until now, the long-tailed video classification is under-explored. 

\subsection{Long-tailed visual classification}
Real-world data usually exhibit long-tailed distribution, which causes models to tend to perform exceptionally well on head classes and overfit on tail classes. The long-tailed visual classification task aims to train models on highly class-imbalanced data and makes it test-generalized. Existing long-tailed methods can be categorized into three classes: re-sampling, re-weighting and ensemble models. (1) Re-sampling methods \cite{cui2019class,chawla2002smote,he2008adasyn} aim to artificially balance the imbalanced data. The strategies could be divided into two popular types: over-sampling and under-sampling. Over-sampling \cite{han2005borderline} repeats the samples from tail classes while under-sampling \cite{drumnond2003class} randomly discards the samples from head classes. BBN \cite{zhou2020bbn} develops a unified Bilateral-Branch Network to handle representation learning and classifier learning jointly. The conventional learning branch is responsible for learning the original data distribution, while the re-balancing branch designs a reversed sampler to model the tail classes. Guo \textit{et al.} \cite{guo2021long} propose a visual recognition network collaboratively to perform well on both head and tail class, where each branch does perform its own binary-cross-entropy-based classification loss with learnable logit compensation. (2) Re-weighting methods \cite{huang2016learning,ting2000comparative,zhou2010multi,dong2017class} usually assign large weights for training samples of tail classes and down-weight the head samples in the loss function. The frequent approach is to apply class weights based on the proportional inverse of class frequency \cite{huang2016learning,wang2017learning}, or the square root of the inverse number of samples \cite{mikolov2013distributed}. Cao \textit{et al.} \cite{cao2019learning} point out that minimizing a margin-based generalization bound can replace the standard cross-entropy objective during training and propose a theoretically-principled label-distribution-aware margin (LDAM) loss. EQL \cite{tan2020equalization} explores an equalization loss to ignore gradients for rare categories and solve the problem of long-tailed rare categories recognition. (3) Ensemble learning-based methods \cite{cai2021ace,guo2021long,xiang2020learning} develop multiple experts to solve long-tailed visual learning problems. RIDE \cite{wang2020long} trains multiple experts independently with softmax loss to improve computing efficiency, which can reduce the model variance and the computational cost with a dynamic expert routing module. Wang \textit{et al.} \cite{wang2020devil} propose a two-stage instance segmentation model for long-tailed learning. The original classification stage can maintain the performance on head classes, while the adjusted classification stage aims to alleviate head bias. Some multiple experts approaches have emerged \cite{zhang2021test,li2022trustworthy}. TLC \cite{li2022trustworthy} proposes a multi-expert fusion framework by jointly conducting classification and uncertainty estimation to identify hard samples.
Recently, long-tailed classification task has been developed from the image field to the video field. Zhang \textit{et al.} \cite{zhang2021videolt} explore a simple and effective method for long-tailed video classification. The re-sampling strategy is guided by class distributions derived from knowledge learned by the network. Moreover, they collect a new large-scale and untrimmed long-tailed video classification dataset, VideoLT. We perform our method on this dataset. Different from the above work, we propose a novel multi-expert distribution calibration for long-tailed video classification. Unlike traditional multiple experts work that directly re-samples or re-weights a single sample in an independent expert, we jointly consider intra-class and inter-class distributions to solve the imbalance issue for long-tailed data. 

\subsection{Distribution Learning}
The distribution of each class in the latent space is important for capturing the statistical properties in model learning, especially for the classes with a scarcity of samples. Therefore, some work resorts to distribution learning for visual classification tasks. Kearns \textit{et al.} \cite{kearns1994learnability} first introduce distribution learning, which aims to learn probability distributions from independent draws. They learn distributions generated by simple circuits of parity gates and reduce the distribution learning problem to a related PAC problem. Kalai \textit{et al.} \cite{kalai2010efficiently} propose a polynomial-time algorithm to accurately estimate the mixture parameters based on the data drawn from a mixture of multivariate Gaussians. Dasgupta \textit{et al.} \cite{dasgupta1999learning} learn an unknown mixture of Gaussians with an arbitrary common covariance matrix and arbitrary mixing weights under a provably correct algorithm, which is to reconstruct the original Gaussian centers by the low-dimensional modes. Gao \textit{et al.} \cite{gao2017deep} convert the label of each image into a discrete label distribution and develop a deep label distribution learning (DLDL) method by decoupling the label ambiguity in both feature learning and classifier learning.

Since training models can easily get over-fitted based on biased distributions formed by the scarcity of samples in some classes, there have also been some recent explorations in the few-shot distribution-oriented modeling, which can provide inspiration for solving the shortage of tail class samples to deal with long-tailed distribution problems. Yang \textit{et al.} \cite{yang2020free} assume that each feature representation follows a Gaussian distribution and transfers statistics from the sufficient examples to adjust the distribution of these classes with few samples. ADM \cite{li2021asymmetric} dynamically integrates the relation between the KL divergence measure and the image-to-class measure and calculates their similarity to complete asymmetric distribution relations between a query and a support class. DPGN \cite{yang2020dpgn} jointly considers the instance-level and distribution-level relations and constructs a dual complete graph network to propagate label information from labeled examples to unlabeled examples.

The current long-tailed classification method based on distribution learning focuses on the distribution of tail classes. Samuel \textit{et al.} \cite{samuel2021distributional} point out that the distribution of training samples of tail classes does not represent well the real distribution of the data. They explore a distributional robustness loss for learning a balanced feature extractor. He \textit{et al.} \cite{he2021distilling} train a teacher network from the entire and uniformly distributed dataset and distill knowledge to a student network based on the long-tailed distribution dataset. However, these methods ignore that the test class distribution is different from training data. In this paper, we focus on the long-tailed video classification, modeling the diverse intra-class distribution and inter-class distribution to handle the imbalanced problem.

%% file: our.tex
\begin{figure*}[htbp]
	\centering
	\includegraphics[scale=0.63]{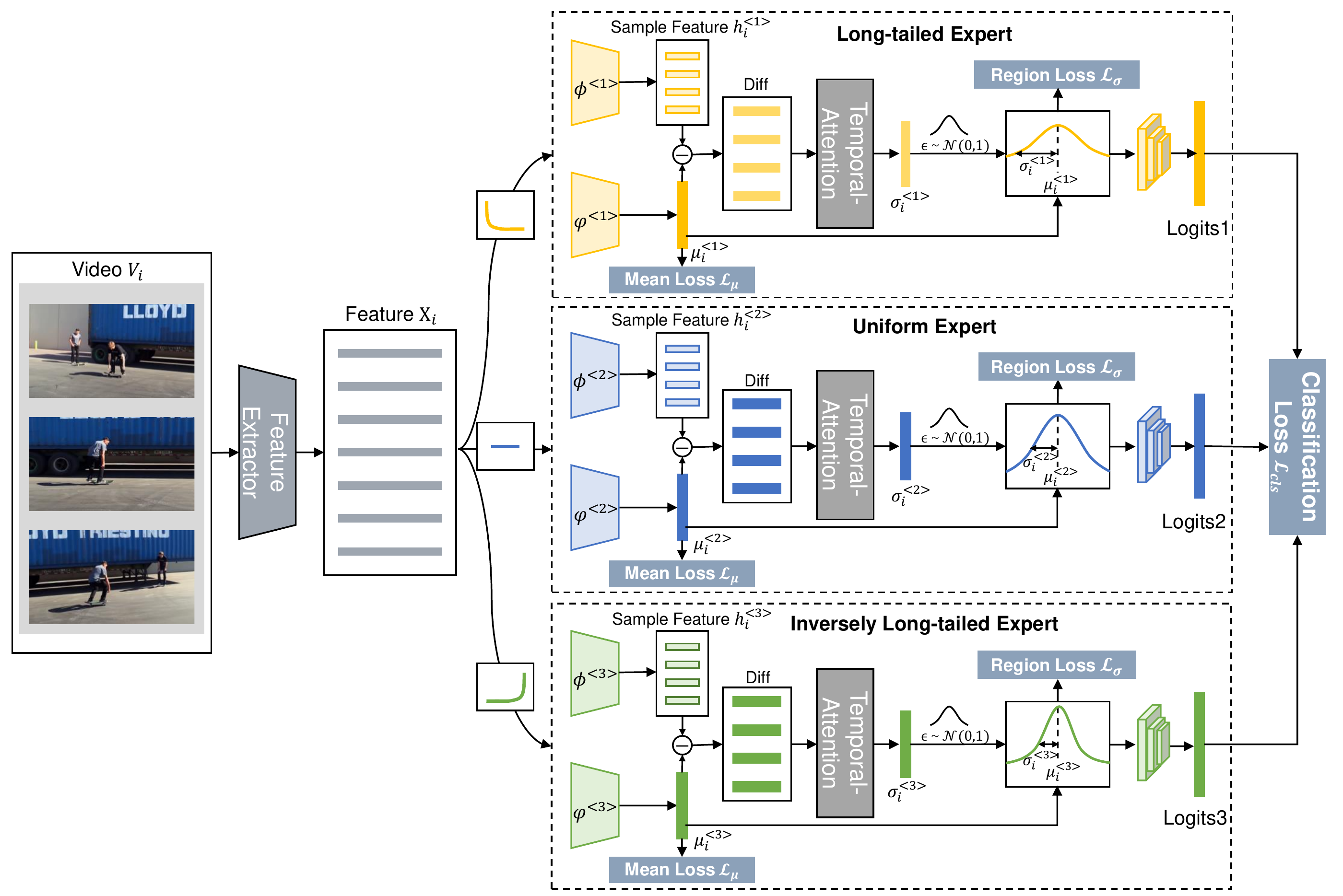}
	\caption{Overview of the proposed multi-expert distribution calibrated method. It considers two-level distribution: (1) We model the parameters mean $\mu$ and variance $\sigma$ of the Gaussian distribution which is designed for estimating the intra-class distribution. (2) Multiple experts are responsible for learning various inter-class distributions: long-tailed expert takes the original data as input, uniform expert takes input data by uniform sampling and inversely long-tailed expert takes inputs by reversing the label frequencies. We take ${\cdot}^{<1>}$, ${\cdot}^{<2>}$ and ${\cdot}^{<3>}$ for representing the three experts.
	} 
	\label{fig:3}
\end{figure*}

In this section, we introduce our proposed multi-expert distribution-calibrated method for long-tailed video classification, which jointly learns the intra-class distribution and the inter-class distribution as shown in Figure \ref{fig:3}. We first show some notations in our framework and then propose the probabilistic formulation and the multi-expert calibration module. Finally, we present the model training and inference process.

\subsection{Notations}
We assume that a long-tailed distribution dataset has $N$ video-label pairs. Each video-label pair denotes as $\left(V_{i}, y_{i}\right), i \in\{1, \cdots N\}$, where $y_{i}=\left[y_{i}^{(1)}, \cdots, y_{i}^{(c)}, \cdots, y_{i }^{(C)}\right]$ is a binary label vector. $y_{i}^{(c)}=1$ indicates the presence of label $c$ in video $i$, while zero otherwise. $C$ is the total number of video categories. For each video $V_{i}$, we extract features $\mathrm{X}_i \in \mathbb{R}^{L \times D}$ by using ResNet-50 or ResNet-101 \cite{he2016deep} pretrained on ImageNet, where $L$ is the number of video frames and $D$ is the feature dimension.

\subsection{Intra-class distribution: Probabilistic Modeling}

In this section, we take advantage of the intra-class distribution, \emph{\ie}, the distribution of samples in a specific class for feature learning. By estimating such distribution information, the representation of a class can be better modeled to combat the bias caused by insufficient and imbalanced data.

\noindent{\textbf{Intra-class Distribution Estimation.}}
To better estimate the intra-class distribution and avoid the over-fitting problem during model training, we assume that the intra-class distribution follows a Gaussian distribution in the latent space, the distribution of samples in the $c$-th class is represented as follows:
\begin{equation}\label{eq.e1}
p\left(\mathrm{X}\mid c\right)=\mathcal{N}\left(\mu^{(c)}, \sigma^{(c)} \mathrm{I}\right),
\end{equation}
where $\mu^{(c)}\in \mathbb{R}^d$ and $\sigma^{(c)}\in \mathbb{R}^d$ are the parameters of the Gaussian distribution, \emph{\ie}, mean and variance, and $\mathrm{I}$ is the identity matrix. Note that other complicate distributions can also be used, such as the mixture of multivariate Gaussians. To avoid complex computation, we estimate the parameters (mean and variance of each class) of the Gaussian distribution through a learnable neural network. As shown in Figure \ref{fig:3}, for the video sample $\mathrm{X}_i$ in class $c$, the mean is predicted by the following function:
\begin{equation}\label{eq.e2}
\mu_{i}^{(c)} = \text{MeanPooling}_T \left( \varphi (\mathrm{X}_i )\right),
\end{equation}
where $\varphi \left(\cdot \right)$ is implemented by a multi-layer perceptron (MLP), which operates on each frame of the video. The $\text{MeanPooling}_T(\cdot)$ operation is applied with a time range $T$. The predicted mean for category $c$, $\mu_{i}^{(c)}$ is a $d$-dimensional vector. In the following, for simplicity, we omit the superscript when there is no ambiguity.

By using Eq.~\ref{eq.e2}, each sample in a class corresponds to different estimated mean values. To accurately estimate the category mean, we adopt a contrastive loss to make the estimated mean of samples from the same class as similar as possible, and the mean of different classes as dissimilar as possible.
\begin{equation}\label{eq.e3}
\mathcal{L}_{\mu}=\log \frac{\exp \left(\mu_i \cdot \mu^+ \right)}{ \exp \left(\mu_i \cdot \mu^+\right)+\sum_{\mu^- \in \mathcal{M}_i}\exp \left(\mu_i \cdot \mu^-\right)}
\end{equation}
where $\mu^+$ and $\mu^-$ are the positive mean and negative mean estimated from same class and different classes, respectively. $\mathcal{M}_i$ is the set of $\mu^-$ from different categories.

To estimate the variance of each category, a simple solution is to calculate the difference between various videos. However, this operation is time-consuming. Note that videos natively contain fruitful and changing temporal information, which can be leveraged for estimating the distribution variation of a specific video category. We propose a temporal attention module to focus on the dependency relations between video frames. With the temporal attention strategy, the information of each frame in the video is directly related to other frames, which is equivalent to a global perception field considering the temporal relations. We then estimate the variance of intra-class distribution based on the output of the temporal attention module:
\begin{equation}\label{eq.e4}
\begin{aligned}
&\alpha_{i l}=\operatorname{softmax}_L\left(\frac{f_{q}\left(h_{i l}-\mu_{i}\right)^{T} \cdot f_{k}\left(h_{i l}-\mu_{i}\right)}{\sqrt{d_{k}}}\right) \\
&\sigma_{i}=\sum_{l=1}^{L} \alpha_{i l} \cdot f_{v}\left(h_{i l}-\mu_{i}\right), h_{i}=\phi\left(\mathrm{X}_{i}\right)
\end{aligned}
\end{equation}
where $\phi\left(\cdot\right)$ operates on each frame of the video implemented by a multilayer perceptron, $h_{i}\in \mathbb{R}^{L\times d}$ and $h_{i l}\in \mathbb{R}^{d}$ are the frame-level features of the $l$-th frame, $\sqrt{d_k}$ is the scaling factor to counteract the extremely small gradients, $f_{q}$, $f_{k}$ and $f_{v}$ are trainable query, key and value mapping functions that map the input to a $d$-dimensional vector, and $\cdot$ denotes the dot product. The attention coefficient $\alpha_{i l}$ is normalized by the softmax function along the time dimension. The normalized attention coefficient $\alpha_{i l}$ and the values mapping function are summed to obtain the variance $\sigma_{i}$ which considers the temporal relations. So far, the representation of each sample in a certain class can be represented as a stochastic embedding sampled from $\mathcal{N}\left(\mu^{(c)}, \sigma^{(c)} \mathrm{I}\right)$.

The stochastic embedding sampling operation is non-differentiable, which prevents the back-propagation of the gradients flow during model training. To enable the probabilistic model with continuous latent variables and make the stochastic embedding sampling operation differentiable, we utilize a re-parameterization strategy \cite{kingma2013auto} using stochastic gradient ascent where gradients are computed by differentiating the lower bound estimator. Based on this strategy, the model can still take the gradient as normal. To perform probabilistic sampling, assuming that $z_{i}$ is a continuous random variable, we use a differentiable transformation of a noise variable $\epsilon$ to make $z_{i}$ as a deterministic variable. $\epsilon$ is an auxiliary noise variable which is sampled from a normal distribution independent of model parameters $\epsilon \sim \mathcal{N}\left(0,1\right)$. Then, we take $z_{i}$ as an equivalent sampling representation and use it for classification.
\begin{equation}\label{eq.e5}
z_{i}=\mu_{i}+\epsilon \sigma_{i}, \epsilon \sim \mathcal{N}\left(0,1\right)
\end{equation}

We take $z_{i}$ as the final representation of the video and as input of a multi-label classifier. The classification loss function is defined as the cross-entropy loss:
\begin{equation}\label{eq.e6}
\begin{aligned}
&p_{i}=\psi_{c l s}\left(z_{i}\right), p_{i}=\left[p_{i}^{(1)}, \cdots, p_{i}^{(c)}, \cdots p_{i}^{(C)}\right] \\
&\mathcal{L}_{c l s}=-\frac{1}{C} \sum_{c=1}^{C} y_{i}^{(c)} \log \left(p_{i}^{(c)}\right)
\end{aligned}
\end{equation}
where $p_{i}^{(c)}$ represents the predicted probability of the sample $z_{i}$ in class $c$.

\subsection{Inter-class distribution: Multi-expert Calibration}
After learning to estimate the intra-class distribution, this section further focuses on the second key insight: when the actual inter-class distribution, \emph{\ie}, global data distribution, cannot be determined, the simulation bias of existing single long-tailed distribution methods will lead to poor generalization of models. In fact, the test data may follow any inter-class distribution. Common distributions are long-tailed, uniform, or inversely long-tailed distributions, \emph{etc}. Most existing long-tailed classification methods attempt to train models to handle test classes with a certain distribution. However, these methods strive to simulate a specific preset class distribution which will suffer from a simulation bias and perform poorly~\cite{zhang2021test}. Therefore, our proposed method utilizes a multi-expert model to simulate various inter-class distributions, making the intra-class distributions estimated by different experts obey the inter-class distribution by distribution calibration.

\noindent{\textbf{Multi-expert Construction Based on Re-sampling.}}
We mainly propose three experts dealing with long-tailed, uniform, and inversely long-tailed distributions. Each expert adopts the first two stages of the ResNet as the expert shared backbone, while the last stage and fully connected layers of the ResNet constitute the independent components of each expert, \emph{\ie}, using different $\phi\left(\cdot \right)$ and $\psi\left(\cdot \right)$ functions to obtain features $z_i$ trained by different experts. Assuming that the number of samples of the $c$-th class is $N^{(c)}$, and the total number of all samples in the training set is $N^{sum}$, the label frequency of class $c$ is defined as $\omega_{c}=\frac{N^{(c)}}{N^{s u m}}$.

Long-tailed distribution expert directly adopts the long-tailed distribution data for model training, maintaining the original distribution of the long-tailed video dataset. Therefore, we perform probabilistic modeling of samples by utilizing the original intra-class distribution. The detailed operation is described in section III-B.

Uniform distribution expert aims to deal with uniformly distributed data, which can be widely applied to class-balanced datasets. We re-sample $z_i$ of each class by using the probabilistic model so that the data exhibit uniform distribution. Specifically, uniform distribution expert uses the prior distribution of label frequencies $\omega^{(c)}$ to adjust the sample distribution so that the occurrence probability of samples in each class is $\frac{1}{C}$.

Similarly, inversely long-tailed expert focuses on inversely long-tailed distribution data. The actual test class distribution in real-world applications may be inversely long-tailed while existing long-tailed methods perform especially poorly in such cases. In this instance, the sampling ratio of each class is proportional to the inverse of the number of samples in each class, \emph{\ie}, the larger sample number in a class, the smaller the sampling ratio needed. We adjust the sample distribution by inverting the prior distribution $\overline{\omega}^{(c)} = resort\left(\omega^{(c)}\right)$ obtained by reversing the order of label frequencies $\omega^{(c)}$.

Note that we utilize the classification loss defined by Eq. \ref{eq.e6} to optimize the three experts, \emph{ie}, $\mathcal{L}_{c l s}=\sum_{i=1}^{3} \mathcal{L}_{c l s}^{<i>}$.

\noindent{\textbf{Inter-class Distribution Calibration.}}
To investigate the distribution more accurately under different experts, we calibrate the region of the intra-class distribution while maintaining the inter-class distribution. As shown in Figure~\ref{fig:3}, for long-tailed distribution, the region (variance) of the head classes will be larger than that of the tail classes. For example, the head class can be set to follow $\sigma^{(c)} \sim \mathcal{N}\left(0,1\right)$ and the tail classes follow $\sigma^{(c)} \sim \mathcal{N}\left(0,0.01\right)$. About uniform distribution, the regions of the head classes and the tail classes are kept as similar as possible, for example, the class distribution follows $\sigma^{(c)} \sim \mathcal{N}\left(0,0.5\right)$. For the inversely long-tailed distribution, we can make the regions of the tail classes larger than that of the head classes, set the head classes to follow $\sigma^{(c)} \sim \mathcal{N}\left(0,0.01\right)$, and the tail classes to follow $\sigma^{(c)} \sim \mathcal{N}\left(0,1\right)$.

We use 2-norm distance to constrain the size of the above class regions for distribution calibration.
\begin{equation}\label{eq.e7}
\mathcal{L}_{\sigma}=\mathbb{E}\left((\sigma^{(c)})^{2}-\gamma^{(c)}\right)
\end{equation}
For uniform distribution classes, $\gamma^{(c)}$ is a constant, while for long-tailed and inversely long-tailed distributions, we perform min-max normalization according to the label frequency to constrain $\gamma^{(c)} \in \left(a,b\right)$.

\subsection{Training and Inference}
For model training, we present an end-to-end unified framework for long-tailed video classification. For all experts, we adopt the mean-estimation objective (mean loss, Eq. \ref{eq.e3}), multi-label cross-entropy objective (classification loss, Eq. \ref{eq.e6}), and distribution calibration loss (region loss, Eq. \ref{eq.e7}) to perform model optimization. The overall loss function of our model is defined as follows:
\begin{equation}\label{eq.e8}
\mathcal{L}=\lambda _1 \mathcal{L}_{\mu}+\lambda _2\mathcal{L}_{c l s}+\lambda _3\mathcal{L}_{\sigma}
\end{equation}
It is worth noting that $\mathcal{L}_{\mu}$, $\mathcal{L}_{c l s}$, and $\mathcal{L}_{\sigma}$ are the aggregation of three experts. $\lambda _1$, $\lambda _2$ and $\lambda _3$ are the loss balance terms.

During inference, we aggregate experts and average the output of their classification results to predict the class of the given test sample and deal with the unknown test class distribution.


%% file: expr.tex
In this section, we evaluate the proposed method on the large-scale long-tailed distribution video dataset VideoLT \cite{zhang2021videolt}. We first provide implementation details and then conduct experiments to demonstrate that our method outperforms state-of-the-art methods. Last, we conduct ablation study and performance analysis that validates the effectiveness of our proposed modules.

\subsection{Dataset}
\noindent{\textbf{VideoLT \cite{zhang2021videolt}}}: The long-tailed distribution dataset contains 1,004 classes and about 256,218 untrimmed videos collected from YouTube, covering a wide range of human activities, including everyday life, housework, sports and travel, \emph{etc}. VideoLT uses $70\%$ videos as training set, $10\%$ videos as validation set and $20\%$ videos as test set. VideoLT defines 47 head classes ($videos> 500$), 617 medium classes ($ 100 <videos\leqslant 500$) and 340 tail classes ($videos\leqslant 100$). Each video may contains multiple labels. 

\subsection{Baselines}
We compare our proposed multi-expert calibration distribution method with several state-of-the-art long-tailed recognition methods as follows:

(1) \noindent{\textbf{LDAM-DRW \cite{cao2019learning}}}: LDAM-DRW proposes a label-distribution-aware loss function to achieve larger margins for tail classes. LDAM loss compares with regularizing by a uniform margin across all classes under the setting of cross-entropy loss and hinge loss. We implement LDAM-DRW by taking the label frequency of each class into account and calculating weights for cross-entropy loss with enforced margins.

(2) \noindent{\textbf{EQL \cite{tan2020equalization}}}: EQL (equalization loss) ignores gradients for rare categories to tackle the problem of long-tailed rare categories. EQL introduces a weight term for each class of each sample, which mainly reduces the influence of negative samples for the tail classes and protects the learning of rare categories from being at a disadvantage during the network parameter updating.

(3) \noindent{\textbf{CBS \cite{kang2019decoupling}}}: CBS (class-balanced sampling) retrains the classifier with class-balanced sampling which is used to alleviate instance-balanced sampling discrepancy. We implement CBS by selecting samples equally from different classes. Specifically, a video is randomly sampled from a certain class, thus videos from both the head and tail classes have equalized sampling possibilities.

(4) \noindent{\textbf{CB Loss \cite{cui2019class}}}: CB Loss explores a theoretical framework to solve data overlap issue and calculates the adequate number of samples in a model and loss-agnostic manner. We implement CB Loss by adding the class-balanced re-weighting term to the loss function, which is inversely proportional to the adequate number of samples.

(5) \noindent{\textbf{Mixup \cite{zhang2018mixup}}}:
Mixup trains a neural network on a convex combination of pairs of samples and their labels, and regularizes the neural network to support simple linear behavior between training examples. Mixup can be implemented in a few lines of code with minimal computational overhead.

(6) \noindent{\textbf{FrameStack \cite{zhang2021videolt}}}: FrameStack proposes a simple and effective long-tailed video classification method by sampling at the frame level to balance the imbalanced class distribution, \emph{\ie}, a plug-in data augmentation strategy to normalize network training, and uses sample-based prediction to adjust binary cross-entropy loss.

\subsection{Implementation Details}
We uniformly sample 150 frames per video as input to the feature extractor and use three popular backbones to extract features of each frame. For the first two backbones, we use ResNet-50 and ResNet-101 \cite{he2016deep} pretrained on ImageNet. We obtain 2048-dimension features resulted from the penultimate layer of the ResNet. Moreover, to keep fairness between the comparison of FrameStack \cite{zhang2021videolt} and our proposed MEDC, we also conduct the experiment with TSM model \cite{lin2019tsm} pretrained on Kinetics-400. For TSM, we utilize ResNet-50 \cite{he2016deep} as its backbone and adopt a NetVLAD model to aggregate features into 64 clusters. We set the hidden layers to 1024 dimensions. 

We first perform probabilistic modeling of samples by utilizing label frequencies to make samples exhibiting roughly different distributions, \emph{\ie}, long-tailed, uniform or inversely long-tailed expert. We then assume that samples in the class exhibit the Gaussian distribution with the parameters $\mu$ and $\sigma$. We adopt a fully-connected layer, a batch normalization layer and an average layer to obtain the mean $\mu$, then subtract the mean from the sample feature, and take the result as input into the temporal attention module to obtain the region size (variance) $\sigma$ of the samples. The temporal attention module is essentially a decoder-encoder. We use the contrastive loss  $\mathcal{L}_{\mu}$ and the region loss function $\mathcal{L}_{\sigma}$ to constrain the intra-class distribution and distribution region size. 

In detail, for uniform distribution, we set $\gamma_{k} = 0.5$ when constraining the region loss. For long-tailed distribution and inversely long-tailed distribution, we set $\gamma_{k} \in \left(0.01,1\right)$ for experts. The final loss $\mathcal{L}_{cls}$ is the arithmetic sum of the classification losses of these experts. We set the initial learning rate of the Adam optimizer to 0.0001, and trains for 120 epochs. The batch size is 128. We implement our method by PyTorch. Experiments are conducted on a Linux server (CPU: Intel(R) Xeon(R) CPU E5-2620 v4@2.60GHz, GPU: NVIDIA TITAN RTX, 256GB RAM).

\noindent{\textbf{Evaluation Metrics}}:
We adopt mAP, Acc@1 and Acc@5 to evaluate the classification performance of our multi-expert distribution calibration method for long-tailed distribution. We evaluate the mean average precision for head classes ($videos> 500$), medium classes ($ 100 <videos\leqslant 500$) and tail classes ($videos\leqslant 100$), respectively. Furthermore, we calculate mAP, Acc@1 and Acc@5 by averaging the classification accuracy of all classes. Our model achieves better performance on the head, medium, and especially tail classes compared with state-of-art methods. 

\subsection{Performance Comparison}
\begin{table*}[!htbp]
\setlength{\tabcolsep}{12.7pt}
\centering
	\caption{Comparisons with the state-of-the-art methods for long-tailed classification using features extracted from ResNet-50, ResNet-101 and TSM (ResNet-50).}
	\label{tab:1}
	\renewcommand\arraystretch{1.3}
	{
\begin{tabular}{lc|cccccc}
\Xhline{1pt}
\multicolumn{2}{c|}{LT-Methods}                      & \multicolumn{1}{c|}{Overall} & \multicolumn{1}{c|}{Head} & \multicolumn{1}{c|}{Medium} & \multicolumn{1}{c|}{Tail} & \multicolumn{1}{c|}{Acc@1} & Acc@5 \\ \Xhline{1pt}
\multicolumn{1}{r|}{\multirow{8}{*}{\rotatebox{90}{ResNet-50}}} & Baseline   & \multicolumn{1}{c|}{0.499}   & \multicolumn{1}{c|}{0.675}         & \multicolumn{1}{c|}{0.553}             & \multicolumn{1}{c|}{0.376}         & \multicolumn{1}{c|}{0.650} & 0.828 \\
\multicolumn{1}{r|}{}& LDAM-DRW   & \multicolumn{1}{c|}{0.502}  & \multicolumn{1}{c|}{0.680}         & \multicolumn{1}{c|}{0.557} & \multicolumn{1}{c|}{0.378}   & \multicolumn{1}{c|}{0.656} & 0.811 \\
\multicolumn{1}{r|}{}& EQL        & \multicolumn{1}{c|}{0.502}   & \multicolumn{1}{c|}{0.679}         & \multicolumn{1}{c|}{0.557} & \multicolumn{1}{c|}{0.378}         & \multicolumn{1}{c|}{0.653} & 0.829 \\
\multicolumn{1}{r|}{}& CBS   & \multicolumn{1}{c|}{0.491}   & \multicolumn{1}{c|}{0.649}         & \multicolumn{1}{c|}{0.545}  & \multicolumn{1}{c|}{0.371}         & \multicolumn{1}{c|}{0.640} & 0.820 \\
\multicolumn{1}{r|}{}& CB Loss    & \multicolumn{1}{c|}{0.495}   & \multicolumn{1}{c|}{0.653}         & \multicolumn{1}{c|}{0.546}             & \multicolumn{1}{c|}{0.381}         & \multicolumn{1}{c|}{0.643} & 0.823 \\
\multicolumn{1}{r|}{} & Mixup      & \multicolumn{1}{c|}{0.484}   & \multicolumn{1}{c|}{0.649}         & \multicolumn{1}{c|}{0.535} & \multicolumn{1}{c|}{0.368}         & \multicolumn{1}{c|}{0.633} & 0.818 \\
\multicolumn{1}{r|}{} & FrameStack & \multicolumn{1}{c|}{0.516}   & \multicolumn{1}{c|}{0.683}         & \multicolumn{1}{c|}{0.569} & \multicolumn{1}{c|}{0.397}         & \multicolumn{1}{c|}{0.658} & 0.834 \\
\multicolumn{1}{r|}{}  & \textbf{MEDC}       & \multicolumn{1}{c|}{\textbf{0.567}}   & \multicolumn{1}{c|}{\textbf{0.720}}         & \multicolumn{1}{c|}{\textbf{0.607}}            & \multicolumn{1}{c|}{\textbf{0.436}}         & \multicolumn{1}{c|}{\textbf{0.667}} & \textbf{0.839} \\ \Xhline{1pt}
\multicolumn{1}{l|}{\multirow{8}{*}{\rotatebox{90}{ResNet-101}}}  & Baseline   & \multicolumn{1}{c|}{0.516}   & \multicolumn{1}{c|}{0.687}         & \multicolumn{1}{c|}{0.568}             & \multicolumn{1}{c|}{0.396}         & \multicolumn{1}{c|}{0.663} & 0.837 \\
\multicolumn{1}{l|}{}                   & LDAM-DRW   & \multicolumn{1}{c|}{0.518}   & \multicolumn{1}{c|}{0.687}         & \multicolumn{1}{c|}{0.572}             & \multicolumn{1}{c|}{0.397}         & \multicolumn{1}{c|}{0.664} & 0.820 \\
\multicolumn{1}{l|}{}                   & EQL        & \multicolumn{1}{c|}{0.518}   & \multicolumn{1}{c|}{0.690}         & \multicolumn{1}{c|}{0.571}             & \multicolumn{1}{c|}{0.398}         & \multicolumn{1}{c|}{0.664} & 0.838 \\
\multicolumn{1}{l|}{}                   & CBS        & \multicolumn{1}{c|}{0.507}   & \multicolumn{1}{c|}{0.660}         & \multicolumn{1}{c|}{0.559}             & \multicolumn{1}{c|}{0.390}         & \multicolumn{1}{c|}{0.652} & 0.828 \\
\multicolumn{1}{l|}{}                   & CB Loss    & \multicolumn{1}{c|}{0.511}   & \multicolumn{1}{c|}{0.665}         & \multicolumn{1}{c|}{0.561}             & \multicolumn{1}{c|}{0.398}         & \multicolumn{1}{c|}{0.656} & 0.832 \\
\multicolumn{1}{l|}{}                   & Mixup      & \multicolumn{1}{c|}{0.495}   & \multicolumn{1}{c|}{0.660}         & \multicolumn{1}{c|}{0.546}             & \multicolumn{1}{c|}{0.381}         & \multicolumn{1}{c|}{0.641} & 0.824 \\
\multicolumn{1}{l|}{}                   & FrameStack & \multicolumn{1}{c|}{0.532}   & \multicolumn{1}{c|}{0.695}         & \multicolumn{1}{c|}{0.584}             & \multicolumn{1}{c|}{0.417}         & \multicolumn{1}{c|}{0.667} & 0.843 \\
\multicolumn{1}{l|}{}                   & \textbf{MEDC}        & \multicolumn{1}{c|}{\textbf{0.603}}   & \multicolumn{1}{c|}{\textbf{0.737}}         & \multicolumn{1}{c|}{\textbf{0.657}}             & \multicolumn{1}{c|}{\textbf{0.499}}         & \multicolumn{1}{c|}{\textbf{0.670}} & \textbf{0.845} \\  \Xhline{1pt}
\multicolumn{1}{l|}{\multirow{8}{*}{\rotatebox{90}{TSM (ResNet-50)}}}  & Baseline   & \multicolumn{1}{c|}{0.565}   & \multicolumn{1}{c|}{0.757}         & \multicolumn{1}{c|}{0.620}             & \multicolumn{1}{c|}{0.436}         & \multicolumn{1}{c|}{0.680} & 0.851 \\
\multicolumn{1}{l|}{}                   & LDAM-DRW   & \multicolumn{1}{c|}{0.565}   & \multicolumn{1}{c|}{0.750}         & \multicolumn{1}{c|}{0.620}             & \multicolumn{1}{c|}{0.439}         & \multicolumn{1}{c|}{0.679} & 0.834 \\
\multicolumn{1}{l|}{}                   & EQL        & \multicolumn{1}{c|}{0.567}   & \multicolumn{1}{c|}{0.757}         & \multicolumn{1}{c|}{0.623}             & \multicolumn{1}{c|}{0.439}         & \multicolumn{1}{c|}{0.681} & 0.852 \\
\multicolumn{1}{l|}{}                   & CBS        & \multicolumn{1}{c|}{0.558}   & \multicolumn{1}{c|}{0.733}         & \multicolumn{1}{c|}{0.616}             & \multicolumn{1}{c|}{0.440}         & \multicolumn{1}{c|}{0.663} & 0.845 \\
\multicolumn{1}{l|}{}                   & CB Loss    & \multicolumn{1}{c|}{0.563}   & \multicolumn{1}{c|}{0.744}         & \multicolumn{1}{c|}{0.616}             & \multicolumn{1}{c|}{0.440}         & \multicolumn{1}{c|}{0.674} & 0.848 \\
\multicolumn{1}{l|}{}                   & Mixup      & \multicolumn{1}{c|}{0.548}   & \multicolumn{1}{c|}{0.736}         & \multicolumn{1}{c|}{0.602}             & \multicolumn{1}{c|}{0.425}         & \multicolumn{1}{c|}{0.665} & 0.846 \\
\multicolumn{1}{l|}{}                   & FrameStack & \multicolumn{1}{c|}{0.580}   & \multicolumn{1}{c|}{0.759}         & \multicolumn{1}{c|}{0.632}             & \multicolumn{1}{c|}{0.459}         & \multicolumn{1}{c|}{0.686} & 0.859 \\
\multicolumn{1}{l|}{}                   & \textbf{MEDC}        & \multicolumn{1}{c|}{\textbf{0.631}}   & \multicolumn{1}{c|}{\textbf{0.783}}         & \multicolumn{1}{c|}{\textbf{0.676}}             & \multicolumn{1}{c|}{\textbf{0.530}}         & \multicolumn{1}{c|}{\textbf{0.693}} & \textbf{0.862} \\
\Xhline{1pt}
\end{tabular}}
\end{table*}

\begin{table}[!htbp]
\centering
	\caption{Comparisons with the state-of-the-art methods for long-tailed classification using TSM (ResNet-50) and features aggreated using NetVLAD.}
	\label{tab:2}
	\renewcommand\arraystretch{1.3}
	\setlength{\tabcolsep}{3mm}{
\begin{tabular}{lc|cccc}
\Xhline{1pt}
\multicolumn{2}{c|}{LT-Methods}                      & \multicolumn{1}{c|}{Overall} & \multicolumn{1}{c|}{Head} & \multicolumn{1}{c|}{Medium} & \multicolumn{1}{c}{Tail} \\ \Xhline{1pt}
\multicolumn{1}{r|}{\multirow{8}{*}{\rotatebox{90}{NetVLAD}}} & Baseline   & \multicolumn{1}{c|}{0.660}   & \multicolumn{1}{c|}{0.803}         & \multicolumn{1}{c|}{0.708}             & \multicolumn{1}{c}{0.554} \\
\multicolumn{1}{r|}{}& LDAM-DRW   & \multicolumn{1}{c|}{0.627}  & \multicolumn{1}{c|}{0.779}         & \multicolumn{1}{c|}{0.675} & \multicolumn{1}{c}{0.519}\\
\multicolumn{1}{r|}{}& EQL        & \multicolumn{1}{c|}{0.665}   & \multicolumn{1}{c|}{0.808}         & \multicolumn{1}{c|}{0.713} & \multicolumn{1}{c}{0.557}          \\
\multicolumn{1}{r|}{}& CBS   & \multicolumn{1}{c|}{0.662}   & \multicolumn{1}{c|}{0.806}         & \multicolumn{1}{c|}{0.708}  & \multicolumn{1}{c}{0.558}     \\
\multicolumn{1}{r|}{}& CB Loss    & \multicolumn{1}{c|}{0.666}   & \multicolumn{1}{c|}{0.801}         & \multicolumn{1}{c|}{0.712}             & \multicolumn{1}{c}{0.566}      \\
\multicolumn{1}{r|}{} & Mixup      & \multicolumn{1}{c|}{0.659}   & \multicolumn{1}{c|}{0.800}         & \multicolumn{1}{c|}{0.706} & \multicolumn{1}{c}{0.556}          \\
\multicolumn{1}{r|}{} & FrameStack & \multicolumn{1}{c|}{0.667}   & \multicolumn{1}{c|}{0.806}         & \multicolumn{1}{c|}{0.713} & \multicolumn{1}{c}{0.566}          \\
\multicolumn{1}{r|}{}  & \textbf{MEDC}       & \multicolumn{1}{c|}{\textbf{0.670}}   & \multicolumn{1}{c|}{\textbf{0.811}}         & \multicolumn{1}{c|}{\textbf{0.716}}            & \multicolumn{1}{c}{\textbf{0.569}}         \\ 

\Xhline{1pt}
\end{tabular}}
\end{table}

\begin{table*}[!htbp]
\centering
	\caption{Ablation experiments of MEDC on the VideoLT dataset.}
	\label{tab:3}
	\renewcommand\arraystretch{1.3}
	\setlength{\tabcolsep}{5mm}{
\begin{tabular}{lc|cccccc}
\Xhline{1pt}
\multicolumn{2}{c|}{LT-Methods}                & \multicolumn{1}{c|}{Overall} & \multicolumn{1}{c|}{Head} & \multicolumn{1}{c|}{Medium} & \multicolumn{1}{c|}{Tail} & \multicolumn{1}{c|}{Acc@1} & \multicolumn{1}{c}{Acc@5} \\ \Xhline{1pt}
\multicolumn{1}{r|}{\multirow{10}{*}{\rotatebox{90}{ResNet-50}}} & Baseline              & \multicolumn{1}{c|}{0.499}   & \multicolumn{1}{c|}{0.675}         & \multicolumn{1}{c|}{0.553}             & \multicolumn{1}{c|}{0.376}         & \multicolumn{1}{c|}{0.650} & 0.828                     \\
\multicolumn{1}{c|}{}  & No Temporal-Attention & \multicolumn{1}{c|}{0.545}   & \multicolumn{1}{c|}{0.712}         & \multicolumn{1}{c|}{0.589}             & \multicolumn{1}{c|}{0.425}         & \multicolumn{1}{c|}{0.659} & 0.825                     \\
\multicolumn{1}{c|}{}  & Expert1               & \multicolumn{1}{c|}{0.526}   & \multicolumn{1}{c|}{0.683}         & \multicolumn{1}{c|}{0.562}             & \multicolumn{1}{c|}{0.377}         & \multicolumn{1}{c|}{0.649} & 0.812                     \\
\multicolumn{1}{c|}{}  & Expert2               & \multicolumn{1}{c|}{0.517}   & \multicolumn{1}{c|}{0.679}         & \multicolumn{1}{c|}{0.567}             & \multicolumn{1}{c|}{0.370}         & \multicolumn{1}{c|}{0.647} & 0.801                     \\
\multicolumn{1}{c|}{}  & Expert3               & \multicolumn{1}{c|}{0.527}   & \multicolumn{1}{c|}{0.679}         & \multicolumn{1}{c|}{0.561}             & \multicolumn{1}{c|}{0.384}         & \multicolumn{1}{c|}{0.643} & 0.807                     \\
\multicolumn{1}{c|}{}  & E1+E2   & \multicolumn{1}{c|}{0.543}   & \multicolumn{1}{c|}{0.702}         & \multicolumn{1}{c|}{0.586}             & \multicolumn{1}{c|}{0.406}         & \multicolumn{1}{c|}{0.654} & 0.832 \\
\multicolumn{1}{c|}{}  & E1+E3    & \multicolumn{1}{c|}{0.535}   & \multicolumn{1}{c|}{0.683}         & \multicolumn{1}{c|}{0.588}             & \multicolumn{1}{c|}{0.403}         & \multicolumn{1}{c|}{0.658} & 0.830 \\
\multicolumn{1}{c|}{}  & E2+E3   & \multicolumn{1}{c|}{0.540}   & \multicolumn{1}{c|}{0.697}         & \multicolumn{1}{c|}{0.584}             & \multicolumn{1}{c|}{0.407}         & \multicolumn{1}{c|}{0.656} & 0.828 \\
\multicolumn{1}{c|}{}  & \textbf{MEDC}                   & \multicolumn{1}{c|}{\textbf{0.567}}   & \multicolumn{1}{c|}{\textbf{0.720}}        & \multicolumn{1}{c|}{\textbf{0.607}}             & \multicolumn{1}{c|}{\textbf{0.436}}         & \multicolumn{1}{c|}{\textbf{0.667}} & \textbf{0.839}                     \\ \Xhline{1pt}
\multicolumn{1}{c|}{\multirow{10}{*}{\rotatebox{90}{ResNet-101}}}  & Baseline              & \multicolumn{1}{c|}{0.516}   & \multicolumn{1}{c|}{0.687}         & \multicolumn{1}{c|}{0.568}             & \multicolumn{1}{c|}{0.396}         & \multicolumn{1}{c|}{0.663} & 0.837                     \\
\multicolumn{1}{c|}{}  & No Temporal-Attention & \multicolumn{1}{c|}{0.580}   & \multicolumn{1}{c|}{0.709}         & \multicolumn{1}{c|}{0.614}             & \multicolumn{1}{c|}{0.468}         & \multicolumn{1}{c|}{0.668} & 0.837                     \\
\multicolumn{1}{c|}{}  & Expert1               & \multicolumn{1}{c|}{0.546}   & \multicolumn{1}{c|}{0.679}         & \multicolumn{1}{c|}{0.588}             & \multicolumn{1}{c|}{0.461}         & \multicolumn{1}{c|}{0.646} & 0.828                     \\
\multicolumn{1}{c|}{}  & Expert2               & \multicolumn{1}{c|}{0.531}   & \multicolumn{1}{c|}{0.670}         & \multicolumn{1}{c|}{0.577}             & \multicolumn{1}{c|}{0.465}         & \multicolumn{1}{c|}{0.643} & 0.815                     \\
\multicolumn{1}{c|}{}  & Expert3               & \multicolumn{1}{c|}{0.547}   & \multicolumn{1}{c|}{0.676}         & \multicolumn{1}{c|}{0.586}             & \multicolumn{1}{c|}{0.462}         & \multicolumn{1}{c|}{0.649} & 0.826                     \\
\multicolumn{1}{c|}{}  & E1+E2    & \multicolumn{1}{c|}{0.588}   & \multicolumn{1}{c|}{0.689}         & \multicolumn{1}{c|}{0.607}             & \multicolumn{1}{c|}{0.483}         & \multicolumn{1}{c|}{0.665} & 0.842 \\
\multicolumn{1}{c|}{}  & E1+E3               & \multicolumn{1}{c|}{0.576}   & \multicolumn{1}{c|}{0.683}         & \multicolumn{1}{c|}{0.602}             & \multicolumn{1}{c|}{0.475}         & \multicolumn{1}{c|}{0.664} & 0.839 \\
\multicolumn{1}{c|}{}  & E2+E3               & \multicolumn{1}{c|}{0.584}   & \multicolumn{1}{c|}{0.708}         & \multicolumn{1}{c|}{0.612}             & \multicolumn{1}{c|}{0.479}         & \multicolumn{1}{c|}{0.667} & 0.840 \\
\multicolumn{1}{c|}{}  & \textbf{MEDC}                   & \multicolumn{1}{c|}{\textbf{0.603}}   & \multicolumn{1}{c|}{\textbf{0.737}}         & \multicolumn{1}{c|}{\textbf{0.657}}             & \multicolumn{1}{c|}{\textbf{0.499}}         & \multicolumn{1}{c|}{\textbf{0.670}} & \textbf{0.845}                     \\ 
\Xhline{1pt}
\end{tabular}}
\end{table*}

Table \ref{tab:1} summarizes the results and comparisons of our MEDC method and other state-of-the-art methods on the features of VideoLT extracted by ResNet50 and ResNet-101 \cite{he2016deep}. Moreover, we investigate the performance using the features extracted from the TSM model pretrained on Kinetics. We evaluate mAP, Acc@1 and Acc@5 in addition to the mean average precision of head classes, medium classes and tail classes. Based on these evaluation metrics, we observe that our method significantly outperforms recent methods.

When using ResNet-50 features, the data augmentation-based Mixup method performs worse than the baseline method due to the direct data mixing strategy between the features cluttering the labels and making the model training difficult. Compared with Mixup, the re-sampling-based class-balanced sampling (CBS) and class-balanced loss (CB Loss) achieve better results, \emph{\ie}, $49.1\%$ and $49.5\%$, which shows that the re-sampling methods can effectively improve the performance of long-tailed video classification. The re-weighting-based method LDAM-DRW and EQL consider the sampling frequency of each class when calculating the weight of cross-entropy or binary cross-entropy, and achieve better classification accuracy $50.2\%$. We observe that long-tailed image classification methods are not suitable for video classification, hence the FrameStack method designed for long-tailed video classification achieves better performance ($51.6\%$). Our method achieves an overall mAP of $56.7\%$, which gains improvement by $6.8\%$ and $6.5\%$ compared with the baseline method and the best performing image-based methods (\emph{\ie} LDAM-DRW and EQL), respectively. Compared with the best performing video-based method (FrameStack), our method also improves the performance by $5.1\%$, demonstrating the effectiveness of the multi-expert distribution calibration method.

Overall, the pros and cons of various approaches on the ResNet-101 features are similar to those of the ResNet-50 features. Mixup achieves the performance $49.5\%$ worse than the baseline method. From Table \ref{tab:1}, we observe that compared with the model using ResNet50 features, using ResNet-101 features significantly improves the classification accuracy, indicating that the feature extractor's network depth will affect the final classification result. The overall mAP is $7.1\%$ higher than the best performing video-based methods (FrameStack) and $8.5\%$ higher than the best performing image-based methods (\emph{\ie} LDAM-DRW and EQL). 

The performance of the medium and tail classes are significantly worse than that of head classes of all methods using ResNet-50 and ResNet-101 features, which highlights the challenges of video classification task for long-tailed distribution. We observe that CB Loss obtains the best performance among long-tailed image classification methods on tail classes when using ResNet-50 and ResNet-101 features ($38.1\%$ and $39.8\%$), respectively. Beyond that, the long-tailed video classification method FrameStack achieves better performance $39.7\%$ and $41.7\%$. Compared with these methods, our method achieves significant improvements of $43.6\%$ and $49.9\%$ in tail classes while maintaining the advance performance of overall and head classes. Extensive experiments show our MEDC outperforms all competitors and demonstrate the effectiveness of our strategy, especially in tail classes.

We further conduct the experiment on TSM \cite{lin2019tsm} model using ResNet-101 \cite{he2016deep} as its backbone, which can validate the generalization of our method for long-tailed video classification. Compared with the image-pretrained features (ResNet-50 and ResNet-101), we achieve $6.4\%$ and $2.8\%$ improvement. In addition, we expand our MEDC with an advanced feature aggregation strategy in Table \ref{tab:2}. We achieve further improvement by $1.0\%$ and $1.5\%$ compared with the NetVLAD baseline for overall and tail classes on the VideoLT dataset using TSM (ResNet-50). It demonstrates the effectiveness of our MEDC in modeling long-tailed distributions. 

\subsection{Further Remarks}
To further understand how the proposed MEDC method improves mAP performance for long-tailed video classification, we investigate the contributions of different components. Concretely, we study the advantage of temporal-attention module and the advantage of different experts. We present the classification accuracy of model variants trained with different components in Table \ref{tab:3}. This section sets up the following ablation experiments:

\begin{figure}[t]
	\centering
	\includegraphics[scale=0.6]{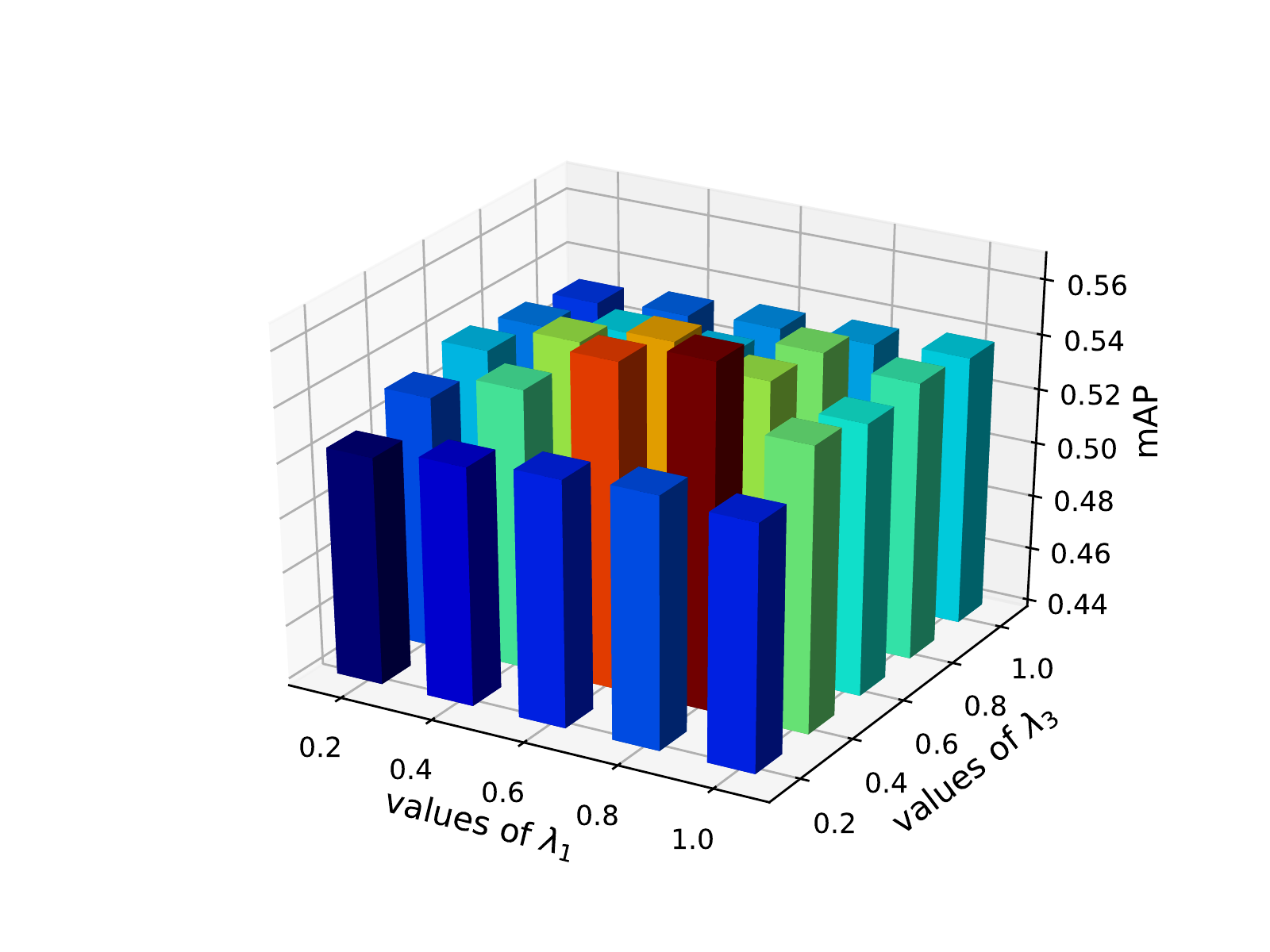}
	\caption{The mAP results on VideoLT dataset with different values of $\lambda_1$ and $\lambda_3$.} 
	\label{fig:4}
\end{figure}

\begin{figure*}[t]
	\centering
	\includegraphics[scale=0.55]{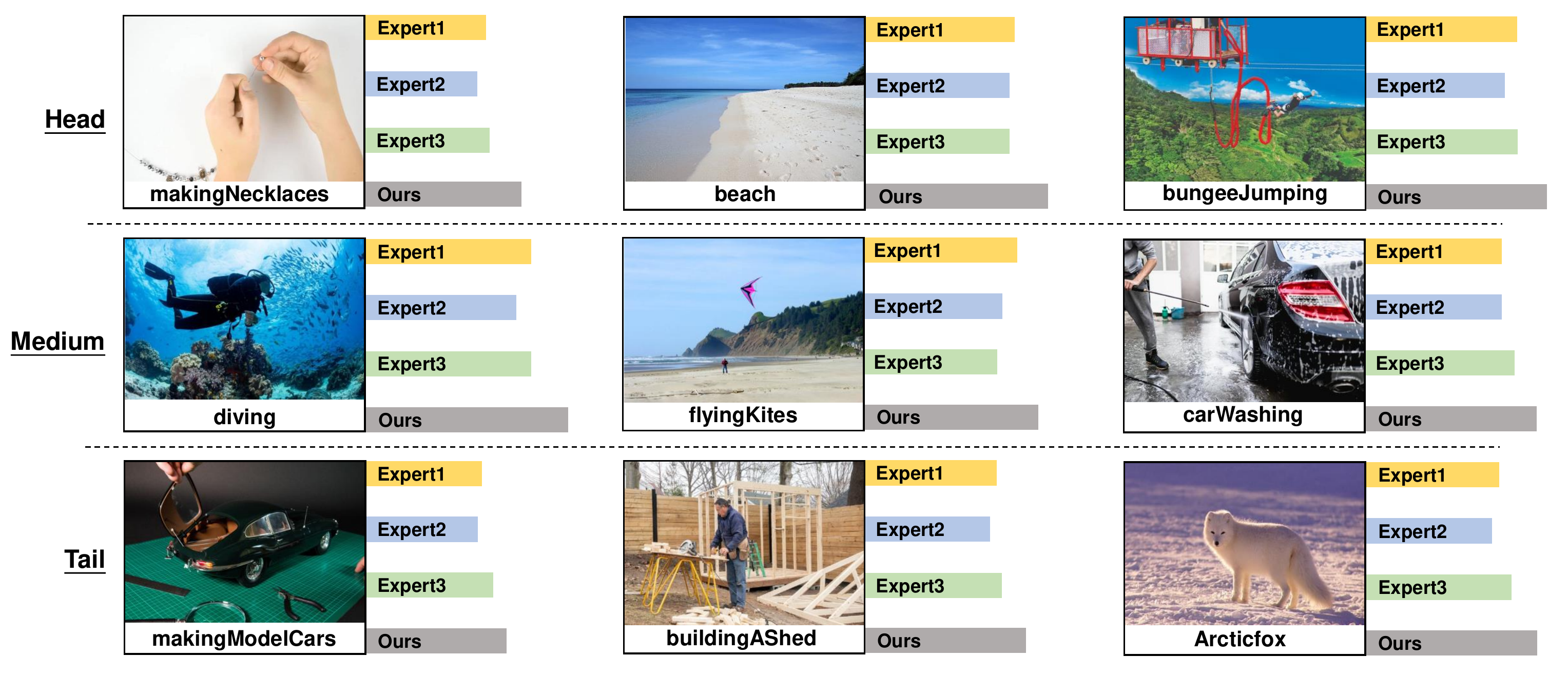}
	\caption{Qualitative results of different experts. We randomly select three classes from the head, medium and tail classes and compare their classification performance of three experts and our MEDC.} 
	\label{fig:5}
\end{figure*}

\textbf{No Temporal-Attention} removes the temporal-attention module. The parameter $\sigma$ of Gaussian distribution is obtained directly using the fully connected layer, the batch normalization layer and the averaging layer.

\textbf{Expert1} uses only long-tailed distribution expert for training. During the re-sampling stage, we maintain that the data are exhibiting in the form of long-tailed distribution, and the other settings remain unchanged.

\textbf{Expert2} means that only uniform distributed expert is used for training, and the settings are as above.

\textbf{Expert3} uses only inversely distributed expert for training, and the settings are as above. 

\begin{figure*}[t]
\centering
\subfigure[Top 10 among 1004 classes that MEDC surpasses FrameStack.]{
		\centering
		    \begin{minipage}[b]{0.45\textwidth}
			\includegraphics[width=\textwidth]{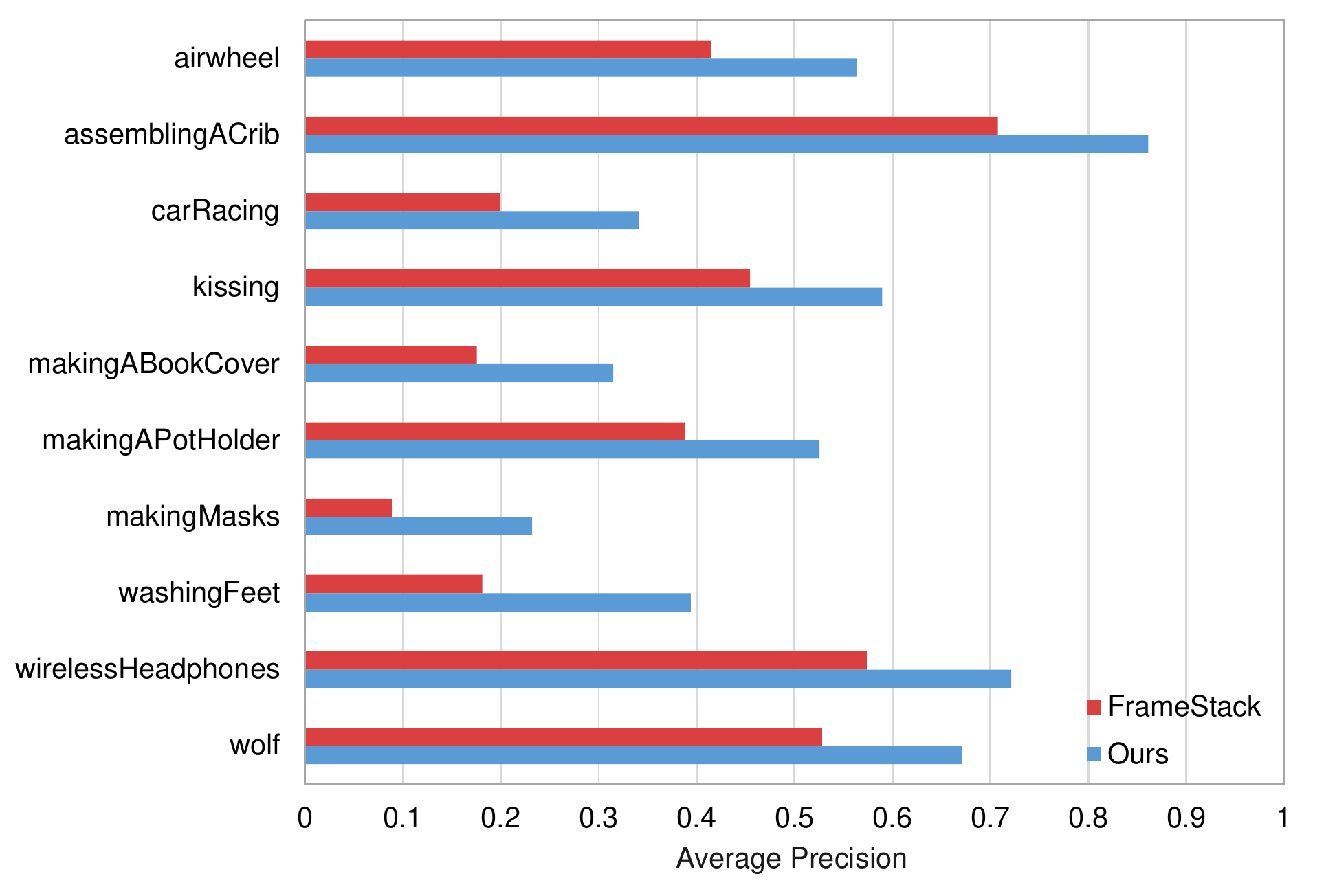} 
		\label{fig:5.1}
				\end{minipage}
		}
    	\subfigure[Top 10 among 340 tail classes that MEDC surpasses FrameStack.]{
    		\begin{minipage}[b]{0.45\textwidth}
   		 	\includegraphics[width=1\textwidth]{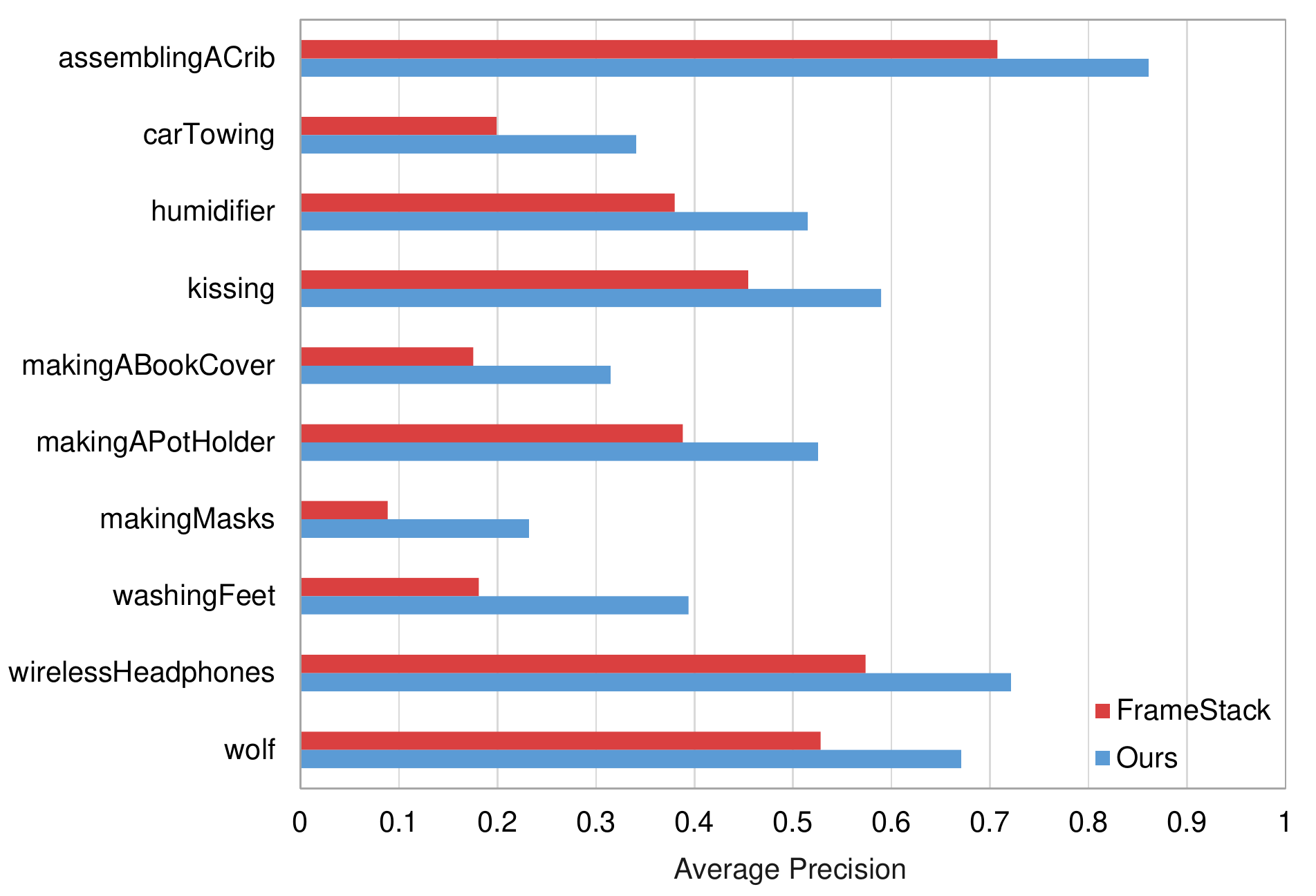}
		\label{fig:5.2}
		    		\end{minipage}
    	}
\\
\subfigure[Comparision on random 10 among 1004 classes.]{
	\begin{minipage}[b]{0.45\textwidth}
	\includegraphics[width=1\textwidth]{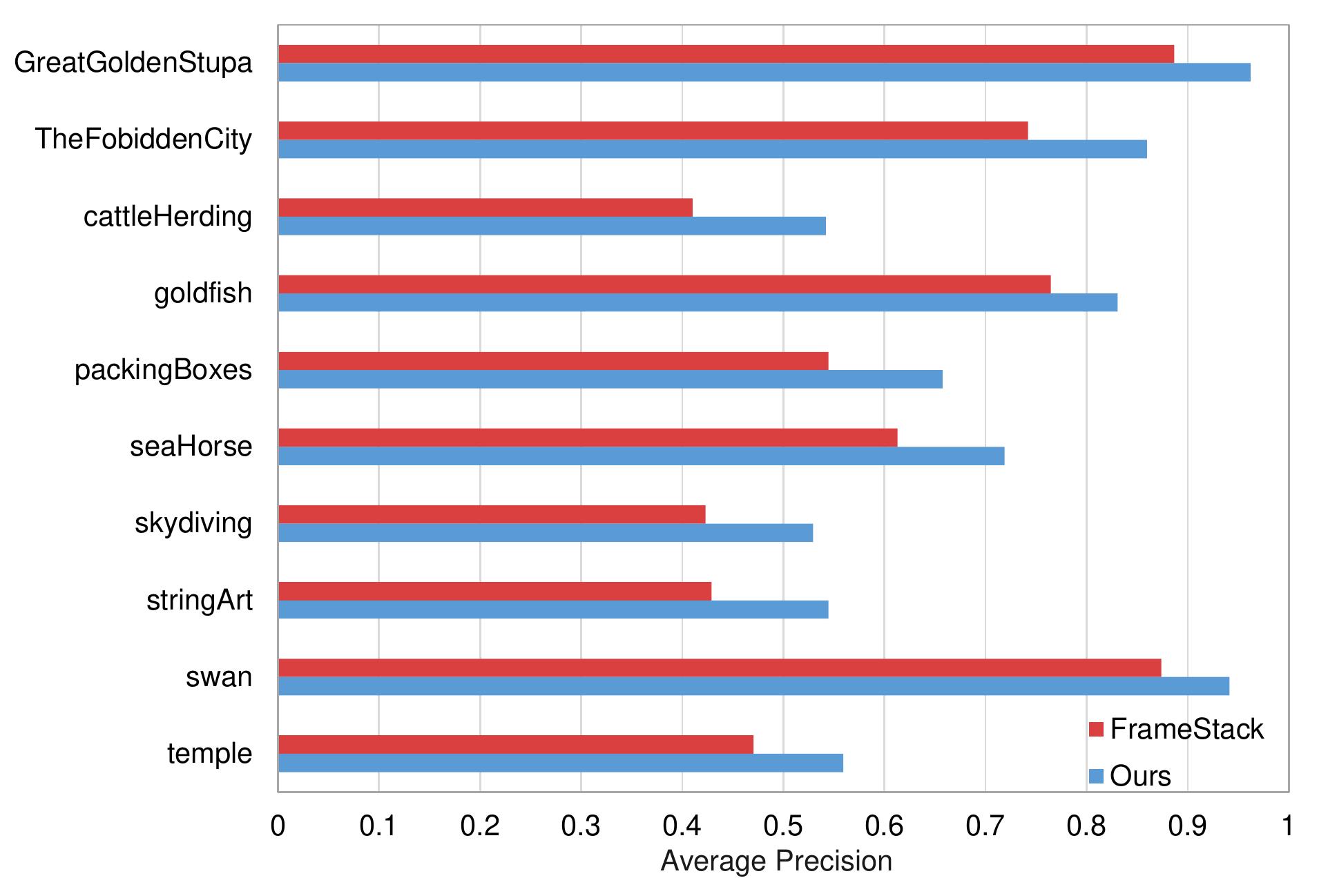} 
	\label{fig:6.1}
	\end{minipage}
	}
\subfigure[Comparision on random 10 among 340 tail classes.]{
    \begin{minipage}[b]{0.45\textwidth}
	\includegraphics[width=1\textwidth]{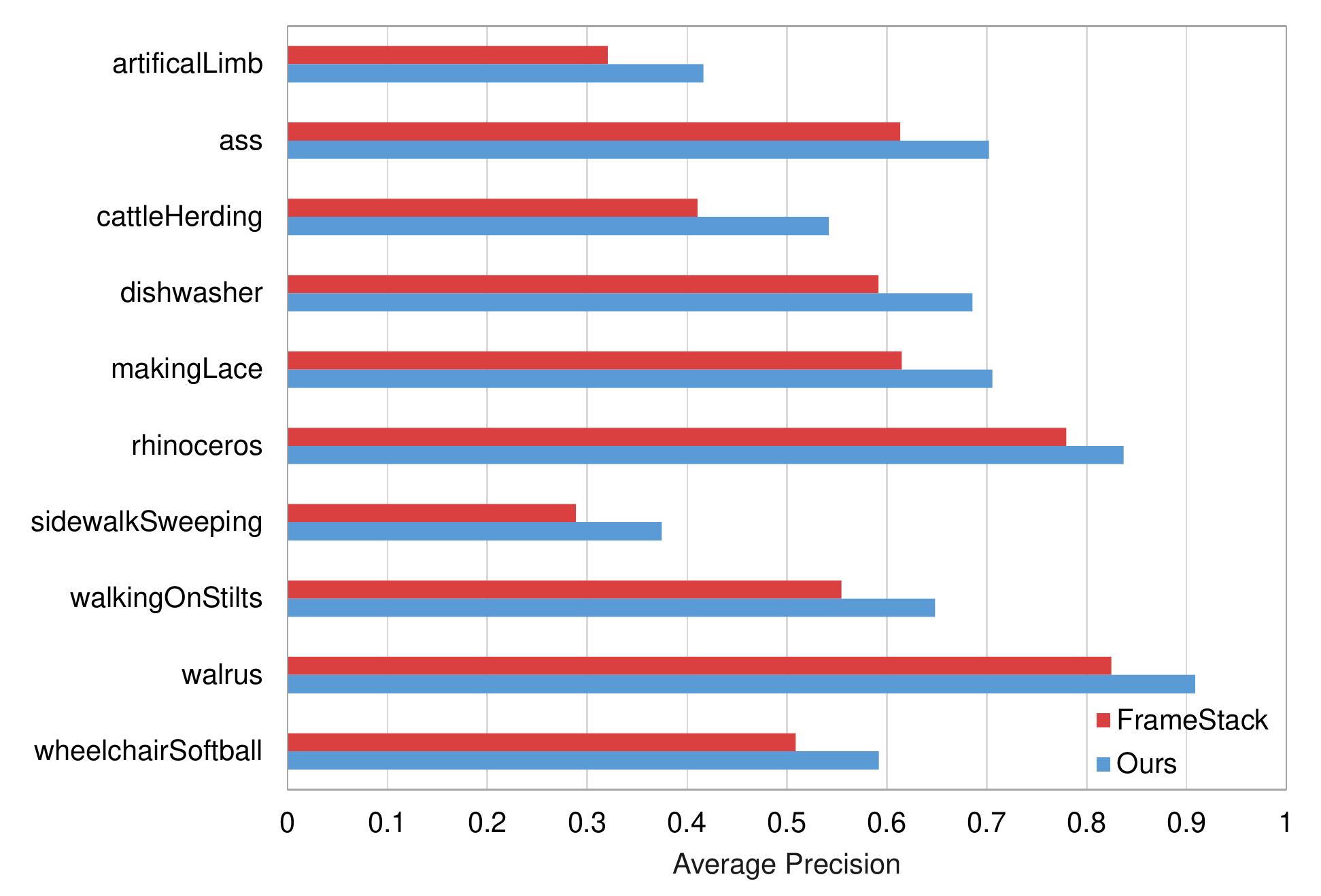}
	\label{fig:6.2}
	\end{minipage}
    	}
	\caption{Performance visualization of FrameStack and our MEDC. We report the top 10 classes that MEDC surpass FrameStack among 1004 classes and 340 tail classes in (a) and (b). We randomly select 10 classes among 1004 classes and 340 tail classes to compare the performance of FrameStack and our MEDC in (c) and (d).}
	\label{fig:56}
\end{figure*}

To demonstrate the effectiveness of the temporal attention module, we use a fully connected layer, a batch normalization layer to obtain the sample variance $\sigma$ directly, and other parameters of Gaussian distribution remain unchanged. We compare the mean average precision of No Temporal-Attention and our MEDC using the ResNet-50 features. The overall mAP performance is improved from $54.5\%$ to $56.7\%$, which shows that capturing the temporal relationship between video frames is beneficial for estimating the distribution region size.

Besides, to understand the effectiveness of the multi-expert model, we compare three different experts independently and keep other modules unchanged, \emph{\ie}, long-tailed distribution expert (E1), uniform distribution expert (E2) and inversely long-tailed distribution expert (E3), respectively. The three experts perform very similarly on various evaluation metrics and have similar overall performance on the VideoLT dataset. We observe that the accuracy of uniform distributed expert are little lower than that of the other two experts, proving that the traditional classification model based on manually-screened uniform datasets performs poorly when applied to the long-tailed distribution dataset ($51.7\%$). In contrast, the long-tailed distribution expert and inversely long-tailed distribution expert achieve better results ($52.6\%$ and $52.7\%$). 

Meanwhile, we also report the results of \textbf{E1+E2}, \textbf{E1+E3} and \textbf{E2+E3}. Such experiment results not only demonstrate that the effectiveness of the single expert or any combination of two experts is worse than our multi-expert model, but also clearly illustrates that it is beneficial to improve the generalization performance when the model considers enough differences of classes to handle inter-class distributions. Furthermore, the performance of various variants of our model using ResNet-101 features is roughly similar to the experimental results of the ResNet-50 features, and achieves a slightly better performance of $2\%$-$4\%$ increment. Table \ref{tab:3} shows the classification accuracy of all model variants. We observe that the performance of our model in each evaluation metric is higher than that of any model variant, which proves that our strategy can effectively improve the classification ability on the long-tailed video classification task.

\textbf{Analysis parameter of $\lambda_1$ and $\lambda_3$.} Furthermore, we conduct a series of experiments on the VideoLT dataset to investigate the sensitivity of the balancing parameters $\lambda_1$ and $\lambda_3$. We vary the effects of $\lambda_1$ and $\lambda_3$ from 0.1 to 1 while keeping other parameter $\lambda_2 = 1$ unchanged in Figure \ref{fig:4}. Experimental results demonstrate that the best mAP results can be obtained when $\lambda_1 = 0.8$ and $\lambda_3 = 0.4$.

\textbf{Qualitative Analysis of different experts.} We visualize the influence of long-tailed distribution expert (Expert1), uniform distributed expert (Expert2), inversely long-tailed expert (Expert3), and multi-expert model (MEDC) in Figure \ref{fig:5}. We randomly select nine classes from the head, medium and tail classes from which we draw the following observations. MEDC outperforms all competitors and consistently improves accuracy on the head, medium and tail classes. Extensive experimental results show that MEDC has the robust classification ability, especially in recognizing tail classes.

\textbf{MEDC vs. FrameStack.} 
We conduct experiments on the overall and tail classes to compare our method MEDC with FrameStack by computing the difference in average precision for each class. As shown in Figure \ref{fig:5.1}, we select the top 10 classes from 1004 classes with the largest difference between MEDC and FrameStack, where $60\%$ of them are action classes. The results show that our MEDC is beneficial for long-tailed video classification, especially for action classes. For Figure \ref{fig:5.2}, we observe that $80\%$ classes are the same, \emph{\ie}, $80\%$ classes are tail classes. The results demonstrate that MEDC is successful in recognizing tail classes and outperforms FrameStack. In addition, we randomly select the top 10 classes from 1004 classes and 340 tail classes as shown in Figure \ref{fig:6.1} and Figure \ref{fig:6.2}. Consistent with Figure \ref{fig:5.1} and  Figure \ref{fig:5.2}, for overall and tail class, our method achieves the better performance on classification accuracy, which again demonstrates the effectiveness of our approach.

%% file: conclus.tex
In this work, targeting at long-tailed video classification, we show the limitation of existing long-tailed classification approaches where the samples in each class may exhibit a specific distribution and test samples may follow any class distribution. Therefore, existing methods may perform overfitting and poorly. To this end, we propose a novel end-to-end multi-expert distribution calibration method for long-tailed video classification, which efficiently solves the data imbalance problem by considering the intra-class and inter-class distribution. We perform probabilistic modeling to estimate the distribution of samples in each class and learn multi-expert to excel at handling diverse distributions of overall data. Experimental results on the VideoLT dataset with long-tailed distribution report that our proposed model significantly outperforms state-of-the-art methods. In the future, category-category relations can also be considered as a constraint in distribution modeling. We will also explore more long-tailed distribution tasks, such as long-tailed action prediction and long-tailed video scene graph generation.
